\documentclass[10pt,twocolumn,letterpaper]{article}

\usepackage{cvpr}
\usepackage{times}
\usepackage{epsfig}
\usepackage{graphicx}
\usepackage{amsmath}
\usepackage{amssymb}
\usepackage{subcaption}
\usepackage{appendix}

\usepackage[pagebackref=true,breaklinks=true,letterpaper=true,colorlinks,bookmarks=false,linkcolor=black]{hyperref}

\cvprfinalcopy 

\hypersetup{
colorlinks=true,
urlcolor=black}

\begin{document}

\title{D3D: Distilled 3D Networks for Video Action Recognition}

\author{Jonathan C. Stroud$^{*\dagger}$\\
{\tt stroud@umich.edu}\\
\and
David A. Ross$^*$\\
{\tt dross@google.com}\\
\and
Chen Sun$^*$\\
{\tt chensun@google.com}\\
\and
Jia Deng$^{*\ddagger}$\\
{\tt jiadeng@cs.princeton.edu}\\
\and
Rahul Sukthankar$^*$\\
{\tt sukthankar@google.com}\\
\and
$^*$Google \qquad $^\dagger$~University of Michigan \qquad $^\ddagger$~Princeton University\\
\and \\
\url{http://www.jonathancstroud.com/d3d}
}

\maketitle
\vspace{-10mm}

\begin{abstract}
   State-of-the-art methods for video action recognition commonly use an ensemble of two networks: the spatial stream, which takes RGB frames as input, and the temporal stream, which takes optical flow as input. In recent work, both of these streams consist of 3D Convolutional Neural Networks, which apply spatiotemporal filters to the video clip before performing classification. Conceptually, the temporal filters should allow the spatial stream to learn motion representations, making the temporal stream redundant. However, we still see significant benefits in action recognition performance by including an entirely separate temporal stream, indicating that the spatial stream is ``missing'' some of the signal captured by the temporal stream. In this work, we first investigate whether motion representations are indeed missing in the spatial stream of 3D CNNs. Second, we demonstrate that these motion representations can be improved by distillation, by tuning the spatial stream to predict the outputs of the temporal stream, effectively combining both models into a single stream. Finally, we show that our Distilled 3D Network (D3D) achieves performance on par with two-stream approaches, using only a single model and with no need to compute optical flow.
\end{abstract}

\vspace{-4mm}

\section{Introduction}

\begin{figure}[t]
    \centering
    \includegraphics[width=0.95\linewidth]{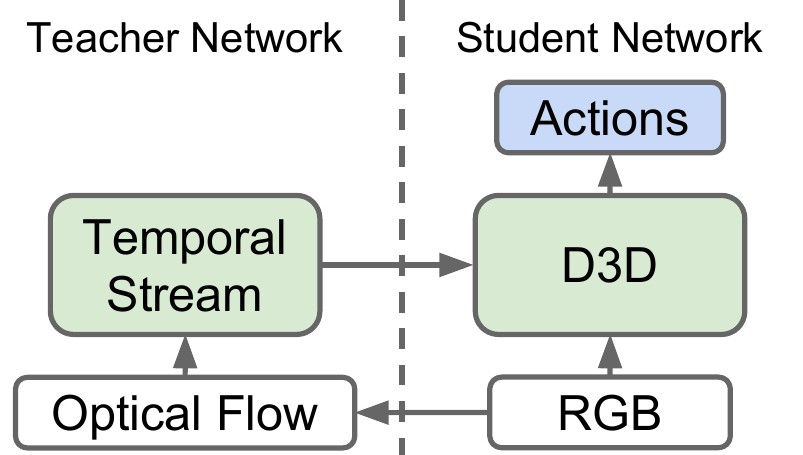}
    \vspace{-2mm}\caption{Distilled 3D Networks (D3D). We train a 3D CNN to recognize actions from RGB video while distilling knowledge from a network that recognizes actions from optical flow sequences. During inference, only D3D is used.}
    \label{fig:d3d}
    \vspace{-3mm}
\end{figure}

Motion is often a necessary cue for recognizing actions in a video clip. For example, it may be difficult to tell two actions apart from a single video frame, like ``open a door'' and ``close a door'', because the interpretation of the action depends on the direction of motion. To handle motion, much recent work on action recognition treats recognition from motion as a task separate from recognition from appearance. Typically, these two tasks are performed by separate networks, the ``temporal stream'' and ``spatial stream'', which are then ensembled, a technique first introduced by Two-Stream Networks~\cite{simonyan2014two}. In Two-Stream Networks, the spatial stream only observes a single RGB video frame at a time, while the temporal stream observes a brief sequence of optical flow frames, meaning the temporal stream is solely responsible for capturing features from motion. However, in more recent work, the spatial stream consists of a 3D Convolutional Neural Network, which observes an entire video clip~\cite{carreira2017quo,xie2018rethinking}. Conceptually, the spatiotemporal filters in a 3D CNN have the ability to respond to movement, which should allow them to learn motion features, a claim echoed in the literature~\cite{tran2015learning,lee2018motion,piergiovanni2018representation}. However, we still see strong gains in accuracy by ensembling these 3D CNNs with ``temporal'' 3D CNNs which take explicit motion representations as input. For example, we see a 6.6\% increase in accuracy on HMDB-51 by ensembling a 3D CNN that takes RGB frames with a 3D CNN that takes optical flow frames~\cite{carreira2017quo}. It is unclear why both streams are necessary. Is the temporal stream capturing motion features which the spatial stream is missing? If so, why is the 3D CNN missing this information? In this work, we examine the spatial streams in 3D CNNs to see what motion representations they learn, and we introduce a method, depicted in Figure~\ref{fig:d3d}, that combines the spatial and temporal streams into a single RGB-only model that achieves comparable performance.

Because 3D CNNs include temporal filters, we hypothesize that they should be able to capture motion representations if optimized to do so. Recent work has shown that it is possible for 3D CNNs to learn motion representations such as optical flow, but the network structure was designed specifically for this purpose~\cite{ng2018actionflownet}. Instead of designing a 3D network specifically for learning motion representations, we examine state-of-the-art 3D CNNs designed for action recognition, with minimal modifications to their structure, to see what motion representations they are capable of learning. To do this, we train 3D CNNs on an optical flow prediction task, described in Section~\ref{predflow}, and we demonstrate experimentally that 3D CNNs are capable of learning motion representations in this way. 

However, even if 3D CNNs are capable of learning motion representations when optimized for optical flow prediction, it is not necessarily true that these motion representations will arise naturally when 3D CNNs are trained to perform other tasks, such as action recognition. To answer whether this is the case, we evaluate the same state-of-the-art 3D CNNs on the optical flow prediction task, but we use models with fixed spatiotemporal filters that are pretrained on an action recognition task. We find that these models underperform models that are fully fine-tuned for optical flow prediction, suggesting that 3D CNNs have much room for improvement to learn higher-quality motion representations.

In order to improve these motion representations, we propose to distill knowledge from the temporal stream into the spatial stream, effectively compressing the two-stream architecture into a single model. In Section~\ref{d3d}, we train this Distilled 3D Network (D3D) by optimizing the spatial stream to match the temporal stream's output, a technique often used for model compression~\cite{hinton2015distilling}. During inference we only use the distilled spatial stream, and we find that this spatial stream has improved performance on the optical flow prediction task. This suggests that distillation improves motion representations in 3D CNNs.

We apply D3D to several benchmark datasets, and we find in Section~\ref{experiments} that D3D strongly outperforms single-stream baselines, achieving accuracy on par with the two-stream model with only a single stream. We train and evaluate D3D on Kinetics~\cite{kay2017kinetics}, and show that the weights learned by distillation also transfer to other tasks, including HMDB-51~\cite{kuehne2011hmdb}, and UCF-101~\cite{soomro2012ucf101}. D3D does not require any optical flow computation during inference, making it less computationally expensive than two-stream approaches. D3D can also benefit from ensembling for better performance, still without the need for optical flow. We compare D3D to a number of other strong RGB-only baselines, and find that D3D outperforms these approaches.

In summary, we make the following contributions:
\vspace{-3mm}
\begin{enumerate}
    \item We investigate whether motion representations arise naturally in the appearance stream of 3D CNNs trained on action recognition.
    \vspace{-3mm}
    \item We introduce a method, Distilled 3D Networks (D3D), for improving these motion representations using knowledge distillation from the temporal stream.
    \vspace{-3mm}
    \item We demonstrate that D3D achieves competitive results on Kinetics, UCF-101, HMDB-51, and AVA, without the need to compute optical flow during inference.
\end{enumerate}

\section{Related Work}

We broadly categorize video action recognition methods into two approaches. First, there are 2D CNN approaches, where single-frame models are used to process each frame individually. Second, there are 3D CNN approaches, where a model learns video-level features using 3D filters. As we will see, both categories of methods often take a two-stream approach, where one stream captures features from appearance, and another stream captures features from motion.

\noindent\textbf{2D CNNs.} Many approaches leverage the strength of single-image (2D) CNNs by applying a CNN to each individual video frame and pooling the predictions across time~\cite{simonyan2014two,donahue2015long,sigurdsson2017asynchronous}. However, na\"\i ve average pooling ignores the temporal dynamics of video. To capture temporal features, Two-Stream Networks introduce a second network called the temporal stream, which takes a sequence of consecutive optical flow frames as input~\cite{simonyan2014two}. The outputs of these networks are then combined by averaging or a linear SVM. Other methods have taken different approaches to incorporating motion by changing the way the features are pooled across time, for example, with an LSTM or CRF~\cite{donahue2015long,sigurdsson2017asynchronous}. These approaches have proven very effective, particularly in the case where video data is limited and therefore training a 3D CNN is challenging. However, recent advances have enabled 3D CNN approaches, which require large video datasets to train, to be effective.

\noindent\textbf{3D CNNs.} Single-frame CNNs can be generalized to video by expanding the filters to three dimensions and applying them temporally, an approach called 3D CNNs. Conceptually, 3D filters should allow CNNs to model motion, but because of the increased number of parameters, 3D CNNs require large amounts of data to train. Large-scale video datasets such as Sports-1M enabled the first 3D CNNs, but these were often not much more accurate than 2D CNNs applied frame-by-frame, calling into question whether 3D CNNs actually model motion~\cite{karpathy2014large}. To compensate, many 3D CNN approaches use additional techniques for incorporating motion. In C3D, motion is incorporated using Improved Dense Trajectory (IDT) features, which leads to a substantial improvement of 5.2\% absolute accuracy on UCF-101~\cite{tran2015learning,wang2013action}. In I3D, S3D-G, and R(2+1)D, using a two-stream approach leads to absolute improvements of 3.1\%,  2.5\%, and 1.1\% on Kinetics, respectively~\cite{carreira2017quo,xie2018rethinking,tran2018closer}. The fact that 3D CNNs benefit from a separate temporal stream suggests that 3D CNNs do not learn to model motion naturally when trained on action recognition tasks. More evidence has shed light on this, for example recent work discovered that 3D CNNs are largely unaffected in accuracy on Kinetics when their input is reversed~\cite{xie2018rethinking}. In addition, it has been shown that using only a single frame from Kinetics videos with C3D achieves only 5\% lower accuracy than using all frames~\cite{huang2018makes}. These results suggest that 3D CNNs do not sufficiently model motion, a hypothesis we explore further in this work.

\noindent\textbf{Why Optical Flow?} If 3D CNNs do not model motion when trained on action recognition, we naturally ask whether motion is even necessary for this task, and if not, what benefits optical flow may offer other than motion. Recent work explored several possible explanations for why optical flow is effective for 3D CNNs~\cite{sevilla2017integration}. One hypothesis is that optical flow is invariant to texture and color, making it difficult to overfit to video datasets when using optical flow representations as inputs. To support this, they demonstrate that action recognition performance is not well correlated with optical flow accuracy, except near motion boundaries and areas of small displacement~\cite{sevilla2017integration}. This work, as well as others, have shown that better or cheaper motion representations can be used in place of optical flow, suggesting that optical flow itself is not necessary for action recognition~\cite{fan2018end,zhang2016real,zhu2017hidden,gao2017im2flow,sevilla2017integration}. Alternatively, optical flow can be used as a source of additional supervision, which is shown by ActionFlowNet~\cite{ng2018actionflownet}, and explored further in this work.

\noindent\textbf{Incorporating Motion in 3D CNNs.} Many approaches have been proposed to incorporate motion features into 3D CNNs without the use of optical flow inputs. Motion Feature Networks, Optical Flow-Guided Features, and Representation Flow all accomplish this by introducing modules into the network architecture which explicitly compute motion representations~\cite{lee2018motion,sun2018optical,piergiovanni2018representation}. Alternatively, several approaches have proposed to replace the optical flow inputs for the temporal stream with a CNN which produces optical flow. For example, Hidden Two-Stream and TVNet use a motion representation that is trained end-to-end for action recognition~\cite{fan2018end,zhu2017hidden}. However, these methods, as well many other methods that propose to use CNNs to predict optical flow, do not use ``vanilla'' 3D CNNs. Instead, they use specialized layers, such as correlations or cost volumes, so they do not answer whether vanilla 3D CNNs can learn motion representations~\cite{ilg2017flownet,sun2018pwc}.

\noindent\textbf{Distillation.} In this work we propose to incorporate motion representations into 3D CNNs using distillation. Distillation was first introduced as a way of transferring knowledge from a teacher network to a (typically smaller) student network by optimizing the student network to reconstruct the output of the teacher network~\cite{hinton2015distilling}. Recent work on distillation has demonstrated that this technique is widely applicable and can be used to transfer knowledge between different tasks or modalities~\cite{furlanello2018born,zhang2016real, qiu2017learning,luo2018graph}. Very related to our work is Motion Vector CNNs, which distill knowledge from the temporal stream into a new motion stream which uses a cheaper motion representation in place of optical flow~\cite{zhang2016real}. Our work differs from this work because we demonstrate that we can distill the temporal stream into the spatial stream and still benefit from its motion representations, without ever explicitly computing motion vectors or evaluating a temporal stream 3D CNN at test time.

\section{Evaluating 3D CNN Motion Representations}

The success of two-stream methods suggests that 3D CNNs do not learn sufficient motion representations on their own. Instead, these methods require that we first estimate optical flow from the RGB frames, and then use it as input to the temporal stream. This avoids the need to learn motion representations, since optical flow already serves as a representation of motion. We ask whether 3D CNNs are capable of learning this representation on their own, which would allow us to eliminate the temporal stream.

We hypothesize that 3D CNNs applied to RGB frames can indeed learn motion representations, but that they do not learn these representations naturally when trained to perform action recognition. To test this, we examine the hidden feature representations of 3D CNNs designed for action recognition. Using these features, we train a decoder to predict optical flow sequences, allowing us to see whether motion representations were captured in these features. We use the performance on the optical flow task as a measure of the architecture's capability to learn motion representations.

\subsection{Predicting Optical Flow}
\label{predflow}

\begin{figure}
    \centering
    \includegraphics[width=0.95\linewidth]{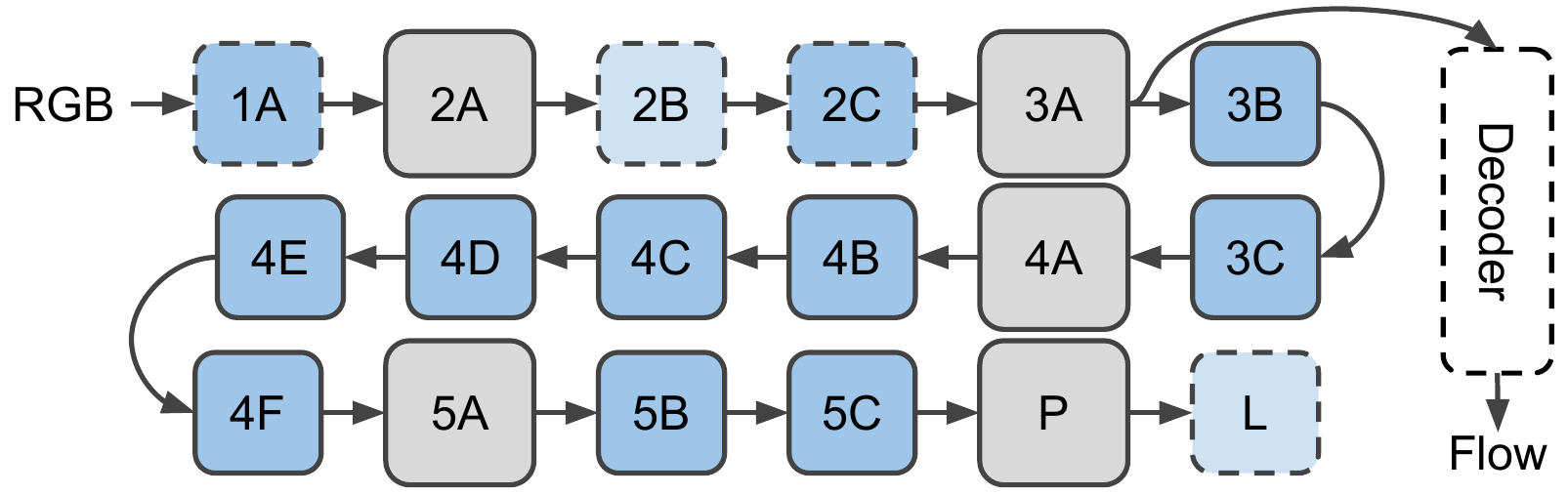}
    \vspace{-2mm}\caption{The network used to predict optical flow from 3D CNN features. We apply the decoder at hidden layers in the 3D CNN (depicted here at layer 3A). This diagram shows the structure of I3D/S3D-G, where blue boxes represent convolution (dashed lines) or Inception blocks (solid lines), and gray boxes represent pooling blocks~\cite{carreira2017quo,xie2018rethinking}. Layer names are the same as those used in Inception~\cite{szegedy2015going}.}
    \label{fig:predflownet}
\end{figure}

\begin{figure}
    \centering
    \includegraphics[width=0.75\linewidth]{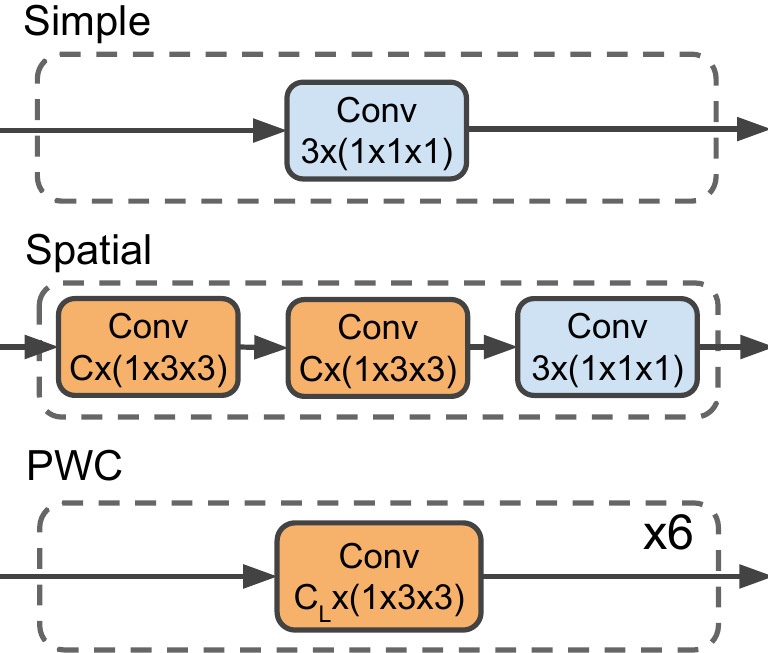}
    \vspace{-2mm}\caption{Three decoders used to predict optical flow. The PWC decoder resembles the optical flow prediction network from PWC-net~\cite{sun2018pwc}. No decoder makes use of temporal filters. See Section~\ref{predflow} for more details.}
    \label{fig:decoders}
    \vspace{-3mm}
\end{figure}

To predict optical flow, we take features from an intermediate layer in a 3D CNN and pass them through a decoder, as depicted in Figure~\ref{fig:predflownet}. For the 3D CNN, we use S3D-G~\cite{xie2018rethinking}. For the decoder, we experiment with three types, which we call ``Simple'', ``Spatial', and ``PWC'', illustrated in Figure~\ref{fig:decoders}. None of these decoders contain temporal filters, cost volume layers, or any other way of incorporating temporal information. This ensures that the decoder is unable to learn motion representations other than what is already represented in the hidden features.

Using the notation $C\times(T\times H \times W)$, we denote the number and size of convolutional filters, where $C$ is the number of output channels, and $T$, $H$, and $W$ describe the size of the kernels in the time, height, and width dimensions. The Simple decoder consists of a single $3\times(1\times1\times1)$ layer. The Spatial decoder consists of two $C\times(1\times3\times3)$ layers, where $C$ is the number of input channels, followed by a $3\times(1\times1\times1)$ layer. The PWC decoder consists of six $C_L\times(1\times3\times3)$ filters, where $C_L$ is the number of channels at layer $L$. For the six layers (from 1 to 6), $C_L=\{128, 128, 96, 64, 32, 3\}$. The PWC decoder is equivalent to the optical flow prediction network in PWC-Net without the cost volume and warping layers~\cite{sun2018pwc}.

The output is a three-channel optical flow representation with the same height and width as the input features. We use the $(\mathit{mag}, \sin{\theta}, \cos{\theta})$ representation used by Im2Flow, where $\mathit{mag}$ and $\theta$ are the magnitude and angle, respectively, of the flow vector at each pixel~\cite{gao2017im2flow}. The decoder is trained to minimize the squared error between this representation and the target optical flow, which is also represented as three channels. For numerical stability, we weight the loss for the $\sin{\theta}, \cos{\theta}$ channels by $\mathit{mag}$.

We use TV-L1 optical flow in place of ground truth optical flow during training~\cite{zach2007duality}. TV-L1 optical flow is commonly used as the input to the temporal stream in many two-stream approaches, indicating that TV-L1 optical flow is useful for action recognition~\cite{carreira2017quo,sevilla2017integration}. Therefore, the ability to reconstruct TV-L1 optical flow with a 3D CNN serves as a measure of how well motion representations for action recognition can be learned with a 3D CNN.

\subsection{Evaluating Motion Representations}
\label{predfloweval}

After training the optical flow decoder, we evaluate the learned optical flow using endpoint error (EPE). The endpoint error is also computed on the estimated TV-L1 optical flow rather than ground truth. In Section~\ref{predflowexperiments}, we evaluate the predicted optical flow at every layer in S3D-G and compare this with that of our proposed method.

We evaluate in two settings. In the first setting, we only train the decoder, and leave the 3D CNN fixed. This settings tests what motion representations are learned by the 3D CNN naturally by training on action recognition. In the second setting, we fine-tune the 3D CNN and decoder end-to-end. This tests what motion representation can be learned by a 3D CNN when optimized specifically for this purpose.

\section{Distilled 3D Networks}
\label{d3d}

Our goal is to incorporate motion representations from the temporal stream into the spatial stream. We approach this using distillation, that is, by optimizing the spatial stream to behave similarly to the temporal stream. Our approach uses the learned temporal stream from the typical two-stream pipeline as a teacher network, and distills the knowledge from this teacher network into a student network, the spatial stream, as depicted in Figure~\ref{fig:d3d}. This is accomplished by introducing a new loss function, which penalizes the outputs of the spatial stream if they are not similar to those of the temporal stream. More concretely, we train the network parameters $\theta$ to minimize the sum of two losses $L_a$ and $L_d$,
\begin{equation}
    L(\theta) = L_a(\theta) + \lambda L_d(\theta)
\end{equation}
where the action classification loss $L_a$ is the cross-entropy and the distillation loss $L_d$ is the mean squared error between the pre-softmax outputs of the spatial stream $f_s(x; \theta)$ and that of the fixed temporal stream $f_t(x)$, i.e.\
\begin{equation}
    L_d(\theta) = \frac{1}{N-1} \sum_{i=0}^N (f_s(x^{(i)}; \theta) - f_t(x^{(i)}))^2,
\end{equation}
where $\{x^{(0)}, ..., x^{(N-1)}\}$ are the video clips. The hyperparameter $\lambda$ allows us to flexibly rescale the contribution of the distillation loss. In our experiments, we find that $\lambda=1$ conveniently serves as a good setting in many cases.

We refer to a spatial stream $f_s$ trained using distillation as Distilled 3D Networks (D3D). During inference, we discard the temporal stream $f_t$, skipping the included optical flow computation step, and rely only on RGB input. As we show in Section~\ref{experiments}, D3D is able to achieve accuracy on par with two-stream methods without the need for two separate spatial and temporal streams. In addition, unlike other approaches for incorporating motion representations, we add no additional computational overhead to the spatial stream~\cite{piergiovanni2018representation,wang2018non,sun2018optical,lee2018motion}. We use S3D-G as the architecture for both D3D and the teacher network, since it achieves comparable accuracy at lower computational cost than competing architectures such as I3D and Non-local I3D~\cite{carreira2017quo,wang2018non}.

\subsection{Alternatives to Distillation}
\label{ablations}

In Section~\ref{ablationexperiments}, we experiment with two alternatives to distillation which underperform D3D. These approaches are described here in detail.

\noindent \textbf{Flow as Supervision.} We introduce an approach similar to ActionFlowNet~\cite{ng2018actionflownet}, which uses TV-L1 optical flow as a source of auxiliary supervision. The network is identical to the method for predicting optical flow with the ``Simple'' decoder at layer ``3A'', described in Section~\ref{predflow}, but we optimize both the 3D CNN and decoder to minimize the sum of the flow prediction loss and the action classification loss. This is a more direct way of encouraging the network to learn motion representations, since we directly penalize the 3D CNN for producing feature representations from which the decoder cannot predict optical flow. However, we find that this does not generally lead to better results on action classification. One possible reason is that the optical flow prediction loss is dominated by the loss applied at background pixels, which covers most of the field of view, whereas prior work demonstrated that accurate optical flow estimation is only correlated with action classification performance near motion boundaries~\cite{sevilla2017integration}. Distillation avoids this because we train the spatial stream to replicate the outputs created by the temporal stream, rather than to directly match its motion representations.
 
\noindent \textbf{Flow as a Learnable Input Representation.} Recent approaches, such as TVNet and Hidden Two-Stream networks, improve upon the temporal stream by learning motion representations specifically for action recognition~\mbox{\cite{zhu2017hidden,fan2018end}.} We introduce an approach similar to these, which uses the first few layers of a 3D CNN as an optical flow prediction network, and passes this predicted flow into a temporal stream, which is all trained end-to-end. We use our optical flow prediction network as in Section~\ref{predflow} up to the ``3A'' layer with the ``Simple'' decoder, and for the temporal stream, we use S3D-G pretrained to predict actions from optical flow. In our experiments, we find that this approach outperforms the temporal stream applied to TV-L1 optical flow, but still underperforms the spatial stream.

\section{Experiments}
\label{experiments}

We train and evaluate D3D on several datasets, demonstrating that D3D outperforms other single-stream models and achieves accuracy on par with that of two-stream models that require explicit optical flow computation.

\subsection{Datasets}
\label{datasets}

\noindent\textbf{Kinetics.} Kinetics is a large-scale video classification dataset with approximately 500K 10-second clips annotated with one of 600 action categories~\cite{kay2017kinetics,carreira2018short}. Kinetics has two variants: Kinetics-600 is the full dataset, and Kinetics-400 is a subset containing 400 of the total categories.

Kinetics consists of publicly available YouTube videos, which can be deleted by their owners at any time. Thus, Kinetics, like similar large-scale Internet datasets, gradually decays over time. Our experiments were conducted in October 2018, when Kinetics-400 contained 226K of the original 247K training examples (-8.4\%) and Kinetics-600 contained 369K of the original 393K training examples \mbox{(-6.1\%)}. The change in both training and validation sets generates a small discrepancy between experiments conducted at different times. We explicitly denote results on the original Kinetics dataset with an asterisk (*) in all tables and provide the list of videos available at the time of our experiments to enable others to reproduce our results \footnote{Code and list of Kinetics videos used are available at the project page: {\tt jonathancstroud.com/d3d}}.

\noindent\textbf{HMDB-51 and UCF-101.} HMDB-51 and UCF-101 are action classification datasets composed of brief video clips, each containing one action~\cite{kuehne2011hmdb,soomro2012ucf101}. HMDB-51 contains 7,000 videos from 51 classes, and UCF-101 contains 13,320 videos from 101 classes. For both datasets, we report classification accuracy on the first test split.

\noindent\textbf{AVA.} AVA is a large-scale spatiotemporal action localization dataset that consists of 430 15-minute movie clips~\cite{gu2017ava}. Each clip contains bounding box annotations at 1-second intervals for all actors in frame, and each actor is annotated with one or more action labels. In our experiments, we train on AVA v2.1, and report results on the validation set.

\subsection{Predicting Optical Flow}
\label{predflowexperiments}

\begin{table}
\begin{center}
\begin{tabular}{l|c|c}
\hline
Decoder & Modality & EPE \\
\hline\hline
All zeros & - & 2.92 \\
PWC & Flow & 0.63 \\ \hline
Simple & RGB & 2.93 \\
Spatial & RGB & 2.34 \\
PWC & RGB & 2.08 \\ \hline
\end{tabular}
\end{center}
\vspace{-5mm}\caption{Effect of decoder on optical flow prediction. We add the optical flow decoder to the ``3A'' layer of S3D-G and train it to predict optical flow.}
\label{tab:predicting-flow}
\vspace{-3mm}
\end{table}

\begin{figure*}[h!]
    \centering
    \includegraphics[width=0.99\linewidth]{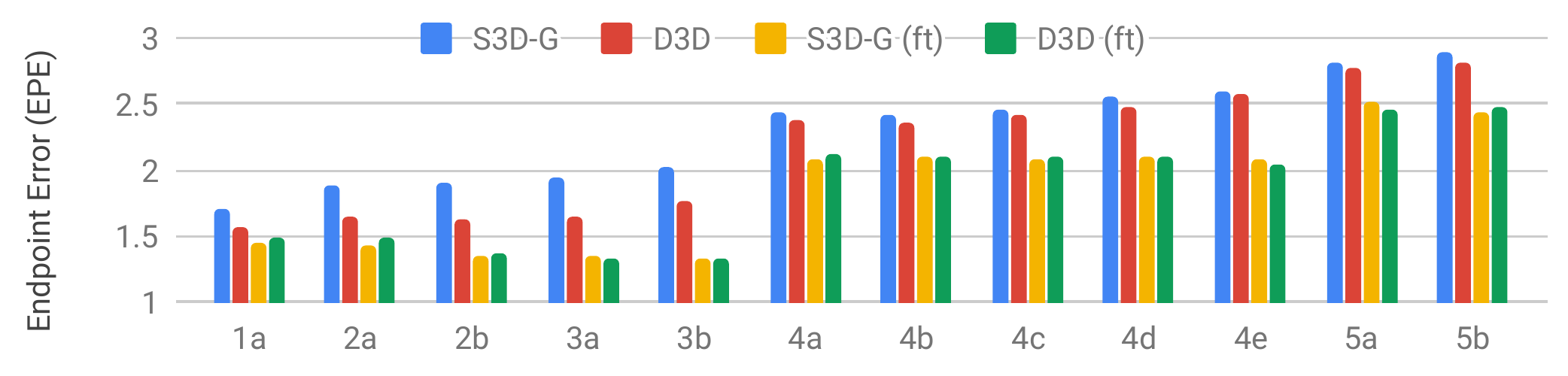}~
    \vspace{-5mm}\caption{Predicting optical flow from each layer in S3D-G and D3D. The horizontal axis indicates which layer (see Figure~\ref{fig:predflownet}) is used as input to the decoder. D3D features are able to more accurately reproduce optical flow across the board, particularly in earlier layers. Fine-tuning either network end-to-end (indicated ``ft''), leads to better performance.}
    \label{fig:predflow}
    \vspace{-3mm}
\end{figure*}

As described in Section~\ref{predfloweval}, we evaluate the ability of 3D CNNs trained for action recognition (specifically on Kinetics-400) to capture motion by attempting to decode optical flow from their hidden feature representations, measuring performance using endpoint error (EPE) compared with the TV-L1 optical flow pseudo-groundtruth. For these experiments, we train each model on 2 GPUs with a batchsize of 6 for 100K iterations, and otherwise we use the same hyperparameters as S3D-G~\cite{xie2018rethinking}.

In Table~\ref{tab:predicting-flow}, we explore the effect of changing the decoder. To bracket performance, we evaluate two baselines: a trivial flow model that predicts ``All zeros'', and a decoder trained on the activations of a temporal stream model, which is provided TV-L1 flow as input. Compared to the baselines, the ``PWC'' decoder trained on spatial stream S3D-G is able approximately estimate optical flow. However, we find that the ``Simple'' decoder is unable to capture any motion signal from the S3D-G features, meaning that motion representations are not readily available from these hidden features.

In Figure~\ref{fig:predflow}, we compare the performance of S3D-G and D3D when the ``PWC'' decoder is applied at each layer. We observed lower error across the board when attempting to predict optical flow from D3D activations versus S3D-G activations. Interestingly, this appears to be strongest in the first few layers, farthest from where the distillation actually takes place. This indicates that distillation is able to improve motion representations even in the earliest layers of 3D CNNs, allowing abstract features to be built from these motion representations in later layers. These results confirm that D3D improves motion representations in 3D CNNs.

\begin{figure}
    \centering
    \begin{minipage}[b]{.5\linewidth}
        \centering\includegraphics[trim={0 0.8cm 0 0.8cm},clip,width=0.75\linewidth]{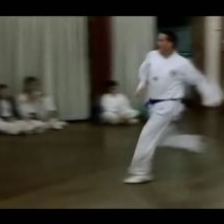}  
        \subcaption{RGB frame}
    \end{minipage}%
    \begin{minipage}[b]{.5\linewidth}
        \centering\includegraphics[trim={0 0.1cm 0 0.1cm},clip,width=0.75\linewidth]{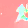}
        \subcaption{TV-L1 optical flow}
    \end{minipage}\\
    \begin{minipage}[b]{.5\linewidth}
        \centering\includegraphics[trim={0 0.1cm 0 0.1cm},clip,width=0.75\linewidth]{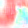}
        \subcaption{S3D-G predicted flow}
    \end{minipage}%
    \begin{minipage}[b]{.5\linewidth}
        \centering\includegraphics[trim={0 0.1cm 0 0.1cm},clip,width=0.75\linewidth]{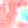}
        \subcaption{D3D predicted flow}
    \end{minipage}
    \vspace{-5mm}\caption{Examples of optical flow produced by S3DG and D3D (without fine-tuning) using the PWC decoder applied at layer 3A. The color and saturation of each pixel corresponds to the angle and magnitude of motion, respectively. TV-L1 optical flow is displayed at $28\times28$px, the output resolution of the decoder.}
    \label{fig:flowexamples}
    \vspace{-3mm}
\end{figure}

When models are fine-tuned end-to-end on the flow prediction task---indicated by (ft) in the figure---the estimates are further improved, and the optical flow performance gap between S3D-G and D3D disappears. When combined with results from Table~\ref{tab:predicting-flow}, these results mostly confirm our original hypothesis: 3D CNNs provided with RGB input have a limited natural tendency to capture the motion signal present in optical flow when trained on action classification. The ability to capture motion signal can be significantly enhanced with modified training objectives, such as distillation loss or by fine-tuning for optical flow prediction.

In Figure~\ref{fig:flowexamples}, we give examples of optical flow estimates given using our method. Both S3D-G and D3D can capture coarse motion, but miss fine details. Results using D3D appear to have slightly more accurate motion boundaries. We provide more examples in the appendix.

\subsection{Distillation on Kinetics}

We train D3D on the Kinetics training set using a two-step procedure. First, we train the teacher S3D-G temporal stream using identical settings to those described in prior work, with TV-L1 optical flow provided by the Kinetics creators~\cite{xie2018rethinking,carreira2017quo}. Second, we train the student D3D network using the distillation procedure described in Section~\ref{d3d}. For fair comparison with S3D-G, we use the same S3D-G hyperparameters when training D3D, with the distillation loss $L_d$ added to the action loss $L_a$ with scaling parameter $\lambda=1$. We train the model for 140k steps on 64 GPUs with a batch size of 6. For details about the hyperparameters, please refer to prior work on S3D-G~\cite{xie2018rethinking}. During inference, the teacher network is not used, optical flow does not need to be computed, and only one network evaluation is performed.

\begin{table}
\begin{center}
\begin{tabular}{l|c|c}
\hline
Method & Modality & Kinetics-400 \\
\hline\hline
ARTNet \cite{wang2017appearance} & RGB+Flow & 72.4* \\
TSN \cite{tran2017convnet} & RGB+Flow & 73.9* \\
R(2+1)D \cite{tran2018closer} & RGB+Flow & 75.4* \\
NL I3D \cite{wang2018non} & RGB & 77.7* \\
SAN \cite{bian2017revisiting} & RGB+Flow+Audio & 77.7* \\ \hline
I3D \cite{carreira2017quo} & RGB & 70.6 / 71.1* \\
I3D \cite{carreira2017quo} & Flow & 62.1 / 63.9* \\
I3D \cite{carreira2017quo} & RGB+Flow & 72.6 / 74.1* \\
S3D-G \cite{xie2018rethinking} & RGB & 74.0 / 74.7* \\
S3D-G \cite{xie2018rethinking} & Flow & 67.3 / 68.0* \\
S3D-G \cite{xie2018rethinking} & RGB+Flow & 76.2 / 77.2* \\\hline
D3D & RGB & 75.9 \\
D3D+S3D-G & RGB+RGB & 76.5 \\\hline
\end{tabular}
\end{center}
\vspace{-5mm}\caption{Performance of D3D on Kinetics-400. All numbers given are top-1 accuracy on the validation set. \mbox{``D3D+S3D-G''} refers to an ensemble of D3D and S3D-G. Numbers marked with an asterisk (*) are reported on the full Kinetics-400 set, those without are reported on the subset available as of October 2018 as described in Section~\ref{datasets}.}
\label{table:kinetics400}
\vspace{-3mm}
\end{table}

In Table~\ref{table:kinetics400}, we compare the performance of D3D with several competitive baselines. We report accuracy for I3D and S3D-G trained and evaluated on the reduced Kinetics-400 dataset described in Section~\ref{datasets}. These replications were run with code provided by the original authors and use identical settings to the published papers. Direct comparison with S3D-G shows that the distillation procedure leads to a 1.9\% improvement in top-1 accuracy, without any additional computational cost during inference. Per-class accuracy is provided in the appendix. Furthermore, we ensemble D3D with S3D-G (``D3D+S3D-G'') by averaging their softmax scores, and achieve a small boost in performance over the two-stream S3D-G approach which uses optical flow. Our ensemble achieves better performance without the need to compute optical flow. The only RGB-only model that outperforms D3D is Non-local I3D, which uses an inflated Resnet-101 backbone architecture with added non-local blocks, which introduce computational overhead during inference~\cite{wang2018non}.

\begin{table}[]
\begin{center}
\begin{tabular}{l|c|c}
\hline
Method & Modality & Kinetics-600 \\
\hline\hline
I3D \cite{carreira2018short} & RGB & 73.6 / 71.9* \\
S3D-G \cite{xie2018rethinking} & RGB & 76.6 \\
S3D-G \cite{xie2018rethinking} & Flow & 69.7 \\
S3D-G \cite{xie2018rethinking} & RGB+Flow & 78.6 \\\hline
D3D & RGB & 77.9 \\
D3D+S3D-G & RGB+RGB & 79.1 \\\hline
\end{tabular}
\end{center}
\vspace{-5mm}\caption{Performance of D3D on Kinetics-600. All numbers given are top-1 accuracy on the validation set. \mbox{``D3D+S3D-G''} refers to an ensemble of D3D and \mbox{S3D-G}. Numbers marked with an asterisk (*) are reported on the full Kinetics-400 set, those without are reported on the subset available as of October 2018 as described in Section~\ref{datasets}. Results on I3D use different settings than in Table~\ref{table:kinetics400}~\cite{carreira2018short}.}
\label{table:kinetics600}
\vspace{-3mm}
\end{table}

In Table~\ref{table:kinetics600}, we compare the performance of D3D with baseline methods on Kinetics-600. Both the teacher and student network are trained using Kinetics-600 in these experiments. We achieve a 1.3\% improvement in single-model performance using D3D, and further improvements by ensembling D3D and S3D-G together, outperforming two-stream S3D-G without the need for optical flow.

\subsection{Transfer to UCF101, HMDB51}

\begin{table}
\begin{center}
\begin{tabular}{l|c|c}
\hline
Method & UCF-101 & HMDB-51 \\
\hline\hline
P3D \cite{qiu2017learning} & 88.6 & - \\
C3D \cite{tran2015learning} & 82.3 & 51.6 \\
Res3D \cite{tran2017convnet} & 85.8 & 54.9 \\
ARTNet \cite{wang2017appearance} & 94.3 & 70.9 \\
I3D \cite{carreira2017quo} & 95.6 & 74.8 \\
R(2+1)D \cite{tran2018closer} & 96.8 & 74.5 \\
S3D-G \cite{xie2018rethinking} & 96.8 & 75.9 \\
I3D Two-Stream \cite{carreira2017quo} & 98.0 & 80.7 \\ \hline
ActionFlowNet \cite{ng2018actionflownet} & 83.9 & 56.4 \\
MFNet \cite{lee2018motion,piergiovanni2018representation} & - & 56.8 \\ 
Rep. Flow \cite{piergiovanni2018representation} & - & 65.4 \\
MV-CNN \cite{zhang2016real} & 86.4 & - \\
TVNet+IDT \cite{fan2018end} & 95.4 & 72.6 \\
Hidden Two-Stream \cite{zhu2017hidden} & 97.1 & 78.7 \\ \hline
D3D (Kinetics-400 pretrain) & 97.0 & 78.7 \\
D3D (Kinetics-600 pretrain) & 97.1 & 79.3 \\
D3D + D3D & 97.6 & 80.5 \\ \hline
\end{tabular}
\end{center}
\vspace{-5mm}\caption{Performance after fine-tuning D3D on UCF-101 and HMDB-51. Our numbers are top-1 accuracy on test split 1 for both datasets. No distillation is performed during fine-tuning.}
\label{table:ucfhmdb}
\vspace{-3mm}
\end{table}

To demonstrate the transferability of D3D, we fine-tune D3D on UCF-101 and HMDB-51. For these experiments, we initialize using D3D pretrained on Kinetics with distillation. However, during fine-tuning, we use only the action classification loss, and not distillation. This avoids the need for a temporal stream altogether, during both training and inference. While we could potentially benefit from applying distillation during fine-tuning as well, these experiments demonstrate that it is not necessary to do so. Each model is fine-tuned for 10k steps on 10 GPUs with a batch size of 6, as described in~\cite{xie2018rethinking}.

In Table~\ref{table:ucfhmdb}, we demonstrate that fine-tuning D3D outperforms many competitive baselines. The models in the top section of the table are strong RGB-only baselines based on 3D CNNs, including S3D-G, which serves as a direct comparison to show that the benefit of distillation during pretraining persists after fine-tuning. The models in the middle section of the table all specifically address the problem of learning motion features without the use of optical flow. D3D outperforms all baselines and achieves essentially equal performance to Hidden Two-Stream when pretrained on Kinetics-400. Hidden Two-Stream uses two I3D models plus an optical flow prediction network, requiring over 2x the cost of D3D for the same accuracy~\cite{zhu2017hidden}.

\subsection{Transfer to AVA}
\label{ava}

\begin{table}
\begin{center}
\begin{tabular}{l|c|c}
\hline
Method & Pretraining & AVA \\
\hline\hline
I3D w/ RPN \cite{girdhar2018better} & Kinetics-600 & 21.9 \\
I3D w/ RPN + JFT \cite{girdhar2018better} & Kinetics-400 & 22.8 \\
S3D-G w/ ResNet RPN \cite{gu2017ava} & Kinetics-400 & 22.0 \\
\hline
D3D w/ ResNet RPN & Kinetics-400 & 23.0 \\ \hline
\end{tabular}
\end{center}
\vspace{-5mm}\caption{Performance on AVA using different backbone networks. All numbers are frame-mAP on the validation set. Models with ``+ ResNet RPN'' use a separate pretrained RPN stream based on ResNet, while the others use the 3D features directly for the RPN. The S3D-G baseline includes changes over the previously published numbers, described in Section~\ref{ava}.}
\label{table:ava}
\vspace{-3mm}
\end{table}

We fine-tune D3D on the spatiotemporal localization dataset AVA. We use a similar approach to the baseline described in the original AVA paper~\cite{gu2017ava}, but adopt some changes introduced by a top entry in the 2018 AVA competition~\cite{girdhar2018better}. Like the original AVA baseline, we use a Faster RCNN-style approach, with a separate pretrained region proposal network (RPN) based on ResNet, and video feature extractor backbone network based on 3D CNNs. Unlike this work, we use D3D in place of I3D as the backbone network. We also adopt the three key changes introduced in the competition entry~\cite{girdhar2018better}. First, we regress only one set of bounding box offsets per region proposal, rather than a different set of offsets per action class. Second, we train for 500k steps using synchronous training on 11 GPUs using a higher learning rate. Third, we add cropping and flipping augmentation during training. Unlike~\cite{girdhar2018better}, we do not remove the ResNet RPN in either D3D our the S3D-G baseline.

In Table~\ref{table:ava}, we compare the use of D3D as a backbone network with S3D-G and I3D. Our approaches use 50 RGB frames and no optical flow. Direct comparison between S3D-G and D3D shows that using D3D leads to a 1\% improvement in Frame-mAP over S3D-G. We also see comparable gains over I3D, and still outperform the I3D-based approach when it includes additional ResNet features pretrained on JFT, an internal Google dataset~\cite{sun2017revisiting}.

\subsection{Ablation study}
\label{ablationexperiments}

\begin{table}
\begin{center}
\begin{tabular}{l|c}
\hline
Method & Kinetics-400 \\
\hline\hline
S3D-G & 74.0 \\
S3D-G temporal stream & 67.3 \\
S3D-G with 3D CNN flow & 69.7 \\
S3D-G with flow loss & 74.3 \\
\hline \hline
D3D distilled at layer 2C & 74.4 \\
D3D distilled at layer 4C & 74.5 \\
D3D distilled at layer 4F & 74.8 \\
\hline \hline
D3D without action loss & 58.8 \\
D3D distilled from spatial stream & 74.3 \\
D3D & 75.9 \\\hline
\end{tabular}
\end{center}
\vspace{-5mm}\caption{Ablation studies. All numbers given are top-1 accuracy on the reduced Kinetics-400 validation set described in Section~\ref{datasets}. D3D using our proposed approach outperforms all other approaches listed. See Section~\ref{ablationexperiments} for details.}
\label{table:ablation}
\vspace{-3mm}
\end{table}

We explore alternative approaches to distillation and how these effect D3D. These results are given in Table~\ref{table:ablation}.

The top section of the table explores non-distillation approaches for improving S3D-G. ``S3D-G temporal stream'' uses TV-L1 optical flow inputs to S3D-G instead of RGB frames, as described in prior work~\cite{xie2018rethinking}. ``S3D-G with 3D CNN flow'' uses optical flow predicted by a separate S3D-G using the approach we introduce in Section~\ref{predflow} instead of TV-L1. The flow prediction network is pretrained to reproduce TV-L1 optical flow, and then is fine-tuned end-to-end with the S3D-G temporal stream on top. This outperforms the TV-L1 optical flow temporal stream, confirming the results of similar end-to-end approaches TVNet and Hidden Two-Stream~\cite{fan2018end,zhu2017hidden}. Finally, ``S3D-G with flow loss'' uses the same flow prediction network as before, but this loss is added to the action loss and both losses are minimized simultaneously as the full S3D-G model is optimized. This leads to slight improvements over S3D-G but does not outperform D3D. Both flow approaches predict at layer ``3A'' and use the ``PWC'' decoder, which we find gives the best results. See Section~\ref{ablations} for more details on these approaches.

The middle section demonstrates applying the distillation loss at intermediate layers. For these experiments, we use $\lambda=1$ except for distillation at the 4f layer, in which case we use $\lambda=100$. We find that we achieve the best performance when setting $\lambda$ such that the scale of the distillation loss roughly matches that of the cross entropy loss. We find that applying the distillation loss at intermediate layers is not as effective as doing so at the network outputs.

The bottom section explores two variants on D3D. ``D3D without action loss'' uses the distillation loss only, and not the cross-entropy loss. ``D3D distilled from spatial stream'' uses the S3D-G spatial stream as the teacher network in place of the temporal stream. Both of these approaches underperform D3D, showing that both losses are useful for D3D, and that distillation alone does not explain the improvement of D3D over S3D-G. Crucially, we only see benefits of distillation when distilling from the temporal stream.

\section{Conclusions}

We introduce D3D, a single-stream distilled 3D CNN which does not require optical flow during inference and still performs on par with two-stream approaches. Furthermore, we show that D3D transfers to other action recognition datasets without the need for further distillation. We study the ability to predict optical flow with 3D CNNs, and we show that 3D CNNs have some limited capacity to learn motion representations, and that D3D reconstructs motion representations better than its non-distilled counterparts.

\section{Acknowledgements}

Work was completed while JCS was an intern at Google Research. We thank our colleagues for their helpful feedback, including Cordelia Schmid, George Toderici, and Carl Vondrick.

{\small
\bibliographystyle{ieee}
\bibliography{d3d}
}

\clearpage
\onecolumn

\appendix
\appendixpage
\section{Predicted Optical Flow Visualizations}

\begin{figure}[h!]
    \centering
    \includegraphics[width=0.5\linewidth]{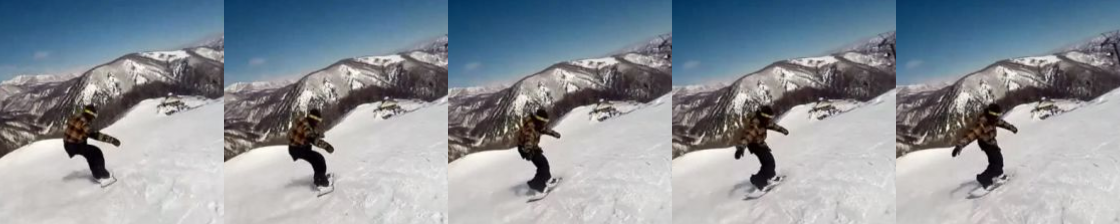}~
    \includegraphics[width=0.5\linewidth]{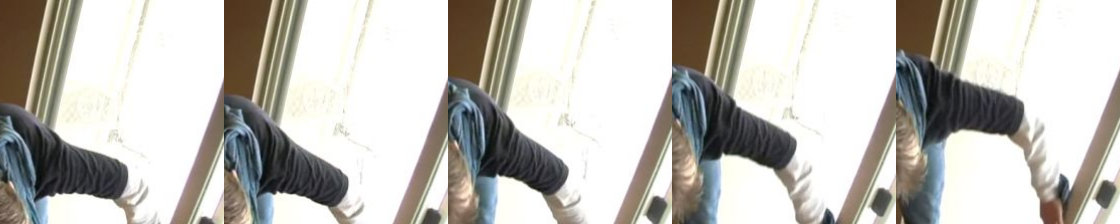}\\
    \includegraphics[width=0.5\linewidth]{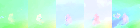}~
    \includegraphics[width=0.5\linewidth]{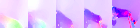}\\
    \includegraphics[width=0.5\linewidth]{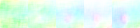}~
    \includegraphics[width=0.5\linewidth]{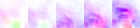}\\
    \includegraphics[width=0.5\linewidth]{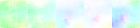}~
    \includegraphics[width=0.5\linewidth]{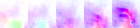}\\
    \includegraphics[width=0.5\linewidth]{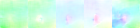}~
    \includegraphics[width=0.5\linewidth]{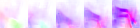}\\
    \caption{Examples of optical flow produced by S3DG and D3D by adding the PWC decoder applied at layer 3A. From top to bottom: RGB Frames, TV-L1 optical flow, S3D-G flow, D3D flow, D3D flow with finetuning. TV-L1 optical flow is shown downsampled to $28\times 28$ px, which is the decoder output resolution used during training.}
    \label{fig:flowsupp}
\end{figure}

\clearpage

\section{Performance on Kinetics-400 Categories}

\begin{figure}[h!]
    \centering
    \includegraphics[width=0.7\linewidth]{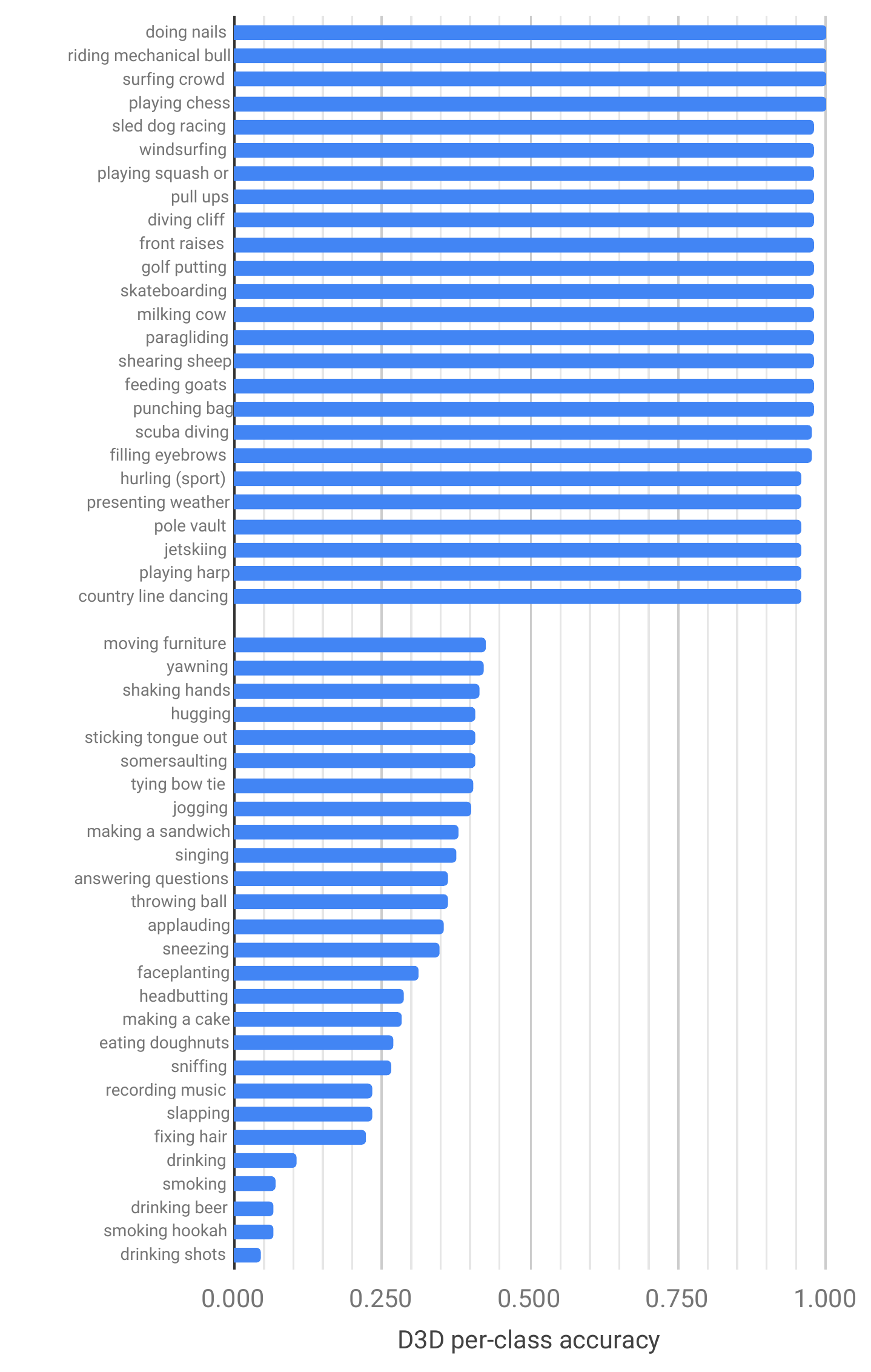}
    \caption{Accuracy on individual Kinetics-400 categories using D3D. We show the per-class accuracy for D3D trained on Kinetics-400. Only the top and bottom 25 classes are shown.}
    \label{fig:kineticssupp}
\end{figure}

\clearpage

\begin{figure}[h!]
    \centering
    \includegraphics[width=0.7\linewidth]{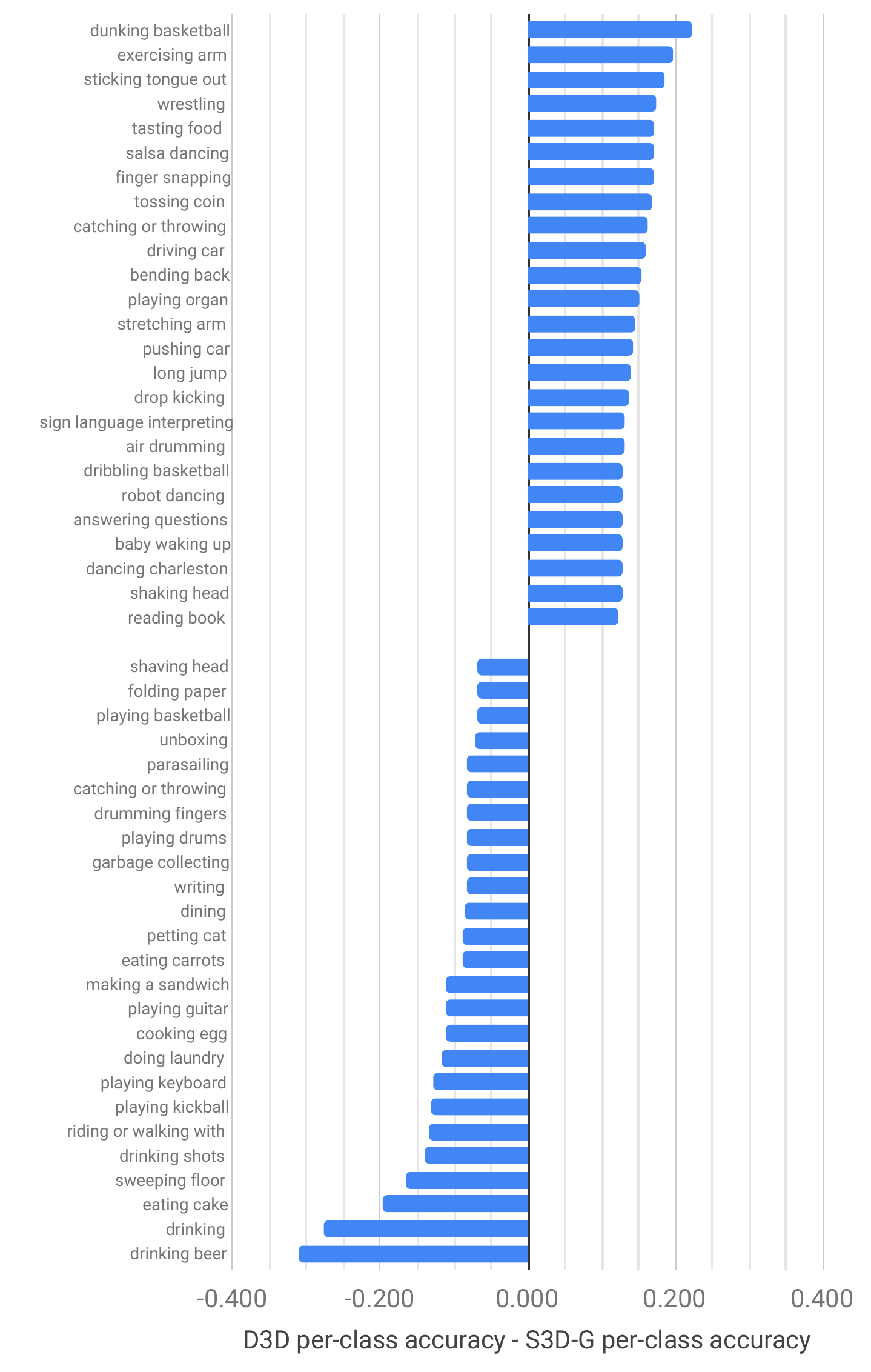}
    \caption{Accuracy difference on individual Kinetics-400 categories by adding distillation. We compare the difference between per-class accuracy for D3D and per-class accuracy for S3D-G. Only the top and bottom 25 classes are shown. In total, D3D leads to improvements on 203 of the 400 classes (50.8\%) and degradations on 103 of the 400 classes (27.3\%), with less than a $\pm.1\%$ difference on the remaining classes.}
    \label{fig:kineticssupp2}
\end{figure}

\end{document}